\documentclass{article}
\usepackage{spconf,amsmath,graphicx}
\usepackage{xcolor}
\usepackage[nolist,nohyperlinks]{acronym}
\usepackage{cleveref}
\usepackage[pdfencoding=auto]{hyperref}
\usepackage{mathtools}
\usepackage{multirow}

\title{CONTEXT-AWARE NEURAL-BASED DIALOG ACT CLASSIFICATION \\ ON AUTOMATICALLY GENERATED TRANSCRIPTIONS}
\name{Daniel Ortega, Chia-Yu Li, Gisela Vallejo, Pavel Denisov, Ngoc Thang Vu}
\address{Institute for Natural Language Processing (IMS), University of Stuttgart, Germany \\
            \{daniel.ortega, thang.vu\}@ims.uni-stuttgart.de}
\begin{document}
%
\maketitle
\begin{abstract}
\end{abstract}
This paper presents our latest investigations on dialog act (DA) classification on automatically generated transcriptions. We propose a novel approach that combines convolutional neural networks (CNNs) and conditional random fields (CRFs) for context modeling in DA classification. We explore the impact of transcriptions generated from different automatic speech recognition systems such as hybrid TDNN/HMM and End-to-End systems on the final performance. Experimental results on two benchmark datasets (MRDA and SwDA) show that the combination CNN and CRF improves consistently the accuracy. Furthermore, they show that although the word error rates are comparable, End-to-End ASR system seems to be more suitable for DA classification.

\section{Introduction}
\label{sec:intro}

According to Austin's theory \cite{Austin1962}, every utterance in a dialog has an illocutionary force, which causes an effect over the course of the conversation. Utterances can then be grouped into \ac{da} categories depending on the relationship between words and the meaning of the expression \cite{Bach1979}. A \ac{da} conveys the intention of the speaker rather than the literal meaning of words for each utterance in a dialog.
	
Automatic \ac{da} classification is a crucial preprocessing step for language understanding and dialog systems. This task has been approached using traditional statistical algorithms, for instance \acp{hmm} \cite{Stolcke2000}, \acp{crf} \cite{Zimmermann2009},  and  more recently  \ac{dl} models, such as \acp{cnn} \cite{Lee2016}, \acp{rnn} \cite{Kalchbrenner2013, Ortega2018} and \ac{am} \cite{Ortega2017, Ortega2018}, achieve state-of-the-art results. 

Several works have shown that context, i.e. preceding utterances, plays an important role at determining automatically the \ac{da} of the current utterance \cite{Lee2016, Ortega2018,  Ortega2017}. 
This fact is also supported by the detailed analysis of the influence of context on \ac{da} recognition presented in \cite{DAcontext:Ribeiro2015}, whose main conclusion is that contextual information helps the \ac{da} classification, as long as such information is distinguishable from the current utterance information. 

In alignment with the aforementioned approaches, we present a model that employs preceding utterances and the current one. However, the particularity of this model relies on using a linear chain \ac{crf} on top of a \ac{cnn} architecture to predict the \ac{da} sequence at utterance level.
Using linear chain \ac{crf} layers on top of \ac{nn} models has been already introduced for sequence labeling tasks at the word level such as named entity recognition \cite{lample2016neural}, part-of-speech tagging \cite{andor2016globally} or for joint entity recognition and relation classification \cite{adel2017global}.

To the best of our knowledge, all work on \ac{da} classification has been done using only \ac{mt}.
Nonetheless, this type of data differs substantially from real data, i.e. \ac{at}, generated by \ac{asr} systems.
In this paper, we explore the effect of training and testing the proposed model on \ac{at}. Our goal at this point is to bring the \ac{da} classification task into a more realistic scenario.

In sum, we introduce a model that combines \acp{cnn} and \acp{crf} for automatic \ac{da} classification. We train and test our model on different scenarios to contrast the effect of using manual and automatically generated transcriptions from two different ASR architectures (hybrid \ac{tdnn}/\ac{hmm} and \ac{e2e} \ac{asr} systems). 
Our results show that the combination of \acp{cnn} and \acp{crf} improves consistently the accuracy of the model achieving state-of-the-art performance on MRDA and SWBD. 
Furthermore, results on ASR outputs reveal that, although \acp{wer} are comparable, the \ac{e2e} \ac{asr} system seems to be more suitable for DA classification.

\vspace{-0.35cm}
\section{Dialog Act Classification}
\label{sec:format}

\begin{figure}[h]
		\begin{minipage}[h]{\linewidth}
			\centering
			\centerline{\includegraphics[width=7cm]{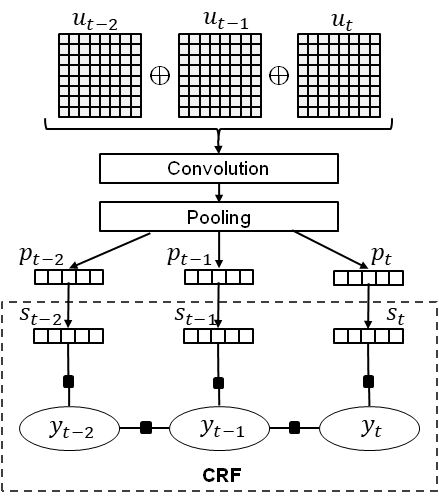}}
		\end{minipage}
		\caption{Model architecture. $\oplus$ stands for concatenation.}
		\label{fig:model}%
	\end{figure}

The \ac{da} classification model proposed in this paper, depicted in \autoref{fig:model}, consists of two parts: a \ac{cnn} that generates vector representations for consecutive utterances and a \ac{crf} that performs \ac{da} sequence labeling. 
	
	\vspace{-1em}
	
	\subsection{Utterance representation}\label{sec:CNN}
	
	Based on \cite{Ortega2017}, the grid-like representations of the current utterance and $n$ previous ones are concatenated and used as input for a \ac{cnn} that generates a vector representation for each of the utterances.
	
	\acp{cnn} perform a discrete convolution using a set of different filters on an input matrix, where each column of the matrix is the word embedding of the corresponding word.
	We use 2D filters $f$ (with width $|f|$) spanning over all embedding dimensions $d$ as described by the following equation:
	\begin{equation}
	(w \ast f)(x,y) = \sum_{i=1}^{d}\sum_{j = -|f|/2}^{|f|/2}w(i,j) \cdot f(x-i,y-j)
	\end{equation}
	
	After convolution, an utterance-wise max pooling operation is applied in order to extract the highest activation. Then, the feature maps are concatenated resulting in one vector per utterance that is represented in Figure \ref{fig:model} as $p_{t-2}, p_{t-1}$ and $p_t$.

	\subsection{\ac{crf}-based \ac{da} sequence labeling}\label{sec:CRF}
	
	Given that a dialog is a sequence of utterances, we approach \ac{da} classification 
	as a sequence labeling problem. Therefore, we employ \acp{crf} for 
	this task. 
	The first step is to generate the score vectors, depicted in Figure \ref{fig:model} as $s_{t-2}, s_{t-1}$ and $s_t$, by the means of a linear function in each time step $t$:
	\begin{equation}
	s_t = Wp_{t}+b
	\end{equation}
	where $W$ (weight matrix) and $b$ (bias) are trainable parameters
	. 
	Using score vectors as input we perform sequence labeling with a \ac{crf} layer. 
	
	\acp{crf} are probabilistic models that calculate the likelihood of a possible output $y$ given an observation $s$. They are commonly represented as factor graphs, in which each factor computes the aforementioned likelihood. Mathematically, each factor graph is defined as:
	\begin{equation}
	p(y|s) = \frac{\prod(\phi(s,y))}{Z(s)}
	\end{equation}
	where $Z(s)$, a normalization function, is the sum of all possible outputs for each observation $s$. 
	
	To perform sequence labeling, we consider a linear chain \ac{crf}. Analogous to Equation 3, the probability of an output sequence $\vec{y}$ given a sequence of observations $\vec{s}$ is:
	\begin{equation}
	p(\vec{y}|\vec{s}) = \frac{\prod(\phi(s,y), \phi'(y,y'))}{Z(\vec{s})}
	\end{equation}
	In Equation 4, not only the factors associating input and output $\phi$ are calculated, but also the likelihood between adjacent labels $\phi'$, where $y$ and $y'$ are neighbors. In this case the normalization function $Z$ takes the sequence $\vec{s}$ as input.

\section{Automatic Speech Recognition}

In recent times, deep learning techniques have boosted the \ac{asr} performance significantly ~\cite{conversational-speech-transcription-using-context-dependent-deep-neural-networks-2}. 
In this section, we introduce the two types of \ac{asr} architectures used to generate \ac{at}.

\subsection{Hybrid \ac{tdnn}/\ac{hmm} architecture}

In hybrid \ac{asr} systems, \acp{nn} are used to predict emission probabilities of \ac{hmm} given speech frames. Recently, various \ac{dl} models have been proposed and developed to improve \ac{asr} performance. Most of them are variations of \acp{cnn} or \acp{rnn} \cite{conversational-speech-transcription-using-context-dependent-deep-neural-networks-2, graves2013speech}. \cite{Povey2016PurelySN} presented a hybrid \ac{tdnn}/\ac{hmm} system trained with lattice-free maximum mutual information, which is fast to train and outperforms significantly other models on many \ac{asr} tasks. To the extent of our knowledge, it is one of the best hybrid \ac{asr} systems available for research and thus was selected for our experiments.
	
\subsection{End-to-End architecture}
More recently, an \ac{e2e} architecture was introduced, which simplifies the training process and achieves competitive results in several benchmark datasets \cite{espnet}. Many studies have proposed \ac{e2e} architectures based on either \ac{ctc} \cite{EESEN} or  \ac{am} \cite{end2endattention}. 

ESPnet, an End-to-End speech processing toolkit, benefits from two major \ac{e2e} ASR implementations based on \ac{ctc} and attention-based encoder-decoder network \cite{espnet}. It employs the multiobjective learning framework to improve robustness and achieve faster convergence. 
For decoding, ESPnet executes a joint decoding by combining both attention-based and \ac{ctc} scores in a one-pass beam search algorithm to eliminate irregular alignments. 
The training loss function is defined in Equation 5, where $\mathcal{L}^{ctc}$ and $\mathcal{L}^{att}$ are the \ac{ctc}-based and attention-based cross entropy, respectively. $\alpha$ is the tuning parameter to linearly interpolate both objective function.
\begin{equation}
    \mathcal{L}=\alpha\mathcal{L}^{ctc} + (1-\alpha)\mathcal{L}^{att}
\end{equation}
During beam search, the following score combination with attention $p^{att}$ and CTC $p^{ctc}$ log probabilities is performed
\begin{equation}
\begin{multlined}
    \texorpdfstring{\log p^{hyb}(y_n|y_{1:n-1}, h_{1:T'}) \\
    = \alpha \log p^{ctc}(y_n|y_{1:n-1}, h_{1:T'}) +\\ (1-\alpha) \log p^{att}(y_n|y_{1:n-1}, h_{1:T'})}{}
\end{multlined}
\end{equation}
where $y_n$ is a hypothesis of output label at position $n$ given a history $y_{1:n-1}$ and encoder output $h_{1:T'}$ \cite{espnet}. 

\section{Experimental setup}
\label{sec:pagestyle}

\subsection{Data for \ac{da} classification}
We evaluate our model on two \ac{da} labeled corpora: 1) \textbf{\acs{icsi}}: \acl{icsi} \cite{Janin2003,Shriberg2004,Dhillon2004}, a dialog corpus of \textit{multiparty meetings}. The 5-tag-set used in this work was introduced by \cite{Ang2005}, and 2) \textbf{\acs{swbd}}: NXT-format Switchboard Corpus \cite{NXT:2010}, a dialog corpus of \textit{2-speaker conversations}.

Train, validation and test splits on both datasets were taken as defined in \cite{Lee2016}\footnote{Concerning \acs{swbd}, the data setup in \cite{Lee2016} was preferred over \cite{Stolcke2000}'s, because it was not clearly found in the latter which conversations belong to each split.}. Table \ref{tab:datasets} presents statistics about the corpora. Both datasets contain a highly unbalanced distribution of classes. The majority class is $59.1$\% on \acs{icsi} and $33.7$\% on \acs{swbd}.
	
\begin{table}[h!]
\centering
		\begin{tabular}{|l|c|c|c|c|c|}
			\hline \textbf{Dataset}& \textbf{C}& \textbf{$\mid$V$\mid$}& \textbf{Train}& \textbf{Validation}&\textbf{Test}\\ \hline
			\acs{icsi}	&5	&12k	& 78k	&16k	&15k\\
			\acs{swbd}	&42	&20k	&193k	&23k	&5k\\
			\hline
		\end{tabular}
	\caption{\label{tab:datasets} Data statistics. \textbf{C}: number of classes, \textbf{$\mid$V$\mid$}: vocabulary size and \textbf{Train}/\textbf{Validation}/\textbf{Test}: number of utterances.}
\end{table}
\vspace{-1em}
		
\subsubsection{Hyperparameters and training}
In Table \ref{tab:hyperparams}, we present the model hyperparameters for both corpora. Most of them were taken from \cite{Ortega2017}. However we tuned the optimizer, the learning rate and the mini-batch size. We found the most effective hyperparameters by changing one at a time while keeping the other ones fixed based on the model performance on the validation split. 

\begin{table}[h]
		    \footnotesize
		    \centering
					\begin{tabular}{|l|c c|}
						\hline 
						\textbf{Hyperparameter}		&\acs{icsi} & \acs{swbd}\\ 
						\hline
						Activation function 		&\multicolumn{2}{c|}{ReLU}  \\
						Dropout rate				&\multicolumn{2}{c|}{0.5}\\
						Filter width 				&\multicolumn{2}{c|}{3, 4, 5} \\
						Filters per width  		    &\multicolumn{2}{c|}{100}  \\
						Learning rate				&0.01 & 0.07 \\
						Mini-batch size				& 70  & 170   \\
						Optimizer					& \acs{gd}  & AdaGrad   \\
						Pooling size		 		&\multicolumn{2}{c|}{utterance-wise}\\
						Word embeddings				&\multicolumn{2}{c|}{word2vec (dim. 300)} \\
						\hline
					\end{tabular}

				
				\caption{Hyperparameters.}
				\label{tab:hyperparams}
\vspace{-5mm}
\end{table}
		
Training was done with context length $n$ ranging from 1-5. After tuning different stochastic learning algorithms with several learning rates, \ac{gd} \cite{Polyak1992} seemed to work best on \acs{icsi} and \ac{adagrad} \cite{Duchi2011} on \acs{swbd}. The learning rate was initialized at 0.01 on \acs{icsi} and 0.07 on \acs{swbd}. Any kind of parameter tuning was done only on the validation split. 
Word vectors were initialized with the 300-dimensional pretrained word vectors from word2vec \cite{Mikolov2013a} and fine-tuned during training. 

\vspace{-1mm}

\subsection{Data for \acl{asr}}
We employed KALDI \cite{Kaldi} to build the hybrid \ac{tdnn}/\ac{hmm} ASR system. In the recipe, 40 \acp{mfcc} were computed at each time step and each frame was append a 100-dimensional iVector to the 40-dimensional \ac{mfcc} input. Speaker adaptive feature transform techniques and data augmentation techniques were implemented. The \ac{gmm}/HMM system generated the alignments for \ac{nn} training \cite{Povey2016PurelySN}.  For the \ac{swbd} dataset, we interpolated the 3-gram language model trained on the transcriptions and the 4-gram Fisher model \cite{fisher}. For \ac{icsi}, we employed a 3-gram language model trained on the \ac{mt}.

\ac{espnet} was used to build the \ac{e2e} \ac{asr} system. The 80-bins log-mel filterbank features with speed-perturbation were used to train the VGG+BLSTM model with five layers encoder with 1024 units and one layer decoder with 1024 units \cite{espnet}. The language model utilized 100 subword units based on byte-pair-encoding technique, which seems to perform better than the character-level language model \cite{Xiao2018HybridCB}. 

Both hybrid \ac{tdnn}/\ac{hmm} and \ac{e2e} \ac{asr} systems were trained on the same train and validation splits and were later used to generate the \acl{at} of all splits (train, validation and test)  for the DA classification model.
Table ~\ref{tab:asr_swbd} shows the performance of hybrid \ac{tdnn}/\ac{hmm} and \ac{e2e} \ac{asr} systems on seen (train and validation splits) data and on unseen data (test split) for \ac{swbd} and \ac{icsi}. 

\begin{table}[h!]
		\centering
		\footnotesize
				\begin{tabular}{|c|c|c|c|c|}
					\hline 
					\multicolumn{1}{|c|}{\multirow{2}{*}{\textbf{Dataset}}}&\textbf{\ac{asr}}& \textbf{Train}& \textbf{Validation}&\textbf{Test}\\ 
					\multicolumn{1}{|c|}{}&\bf System& \textbf{WER}&\textbf{WER}&\textbf{WER}\\ 
					\hline
					\multicolumn{1}{|c|}{\multirow{2}{*}{\ac{swbd}}}&\text{TDNN/HMM}	& 13.8 & 14.29 & 18.02\\
					\multicolumn{1}{|c|}{}&E2E	& 2.90 & 8.90 & 18.80\\
					\hline
					\multicolumn{1}{|c|}{\multirow{2}{*}{\ac{icsi}}}&\text{TDNN/HMM}	& 9.89 & 19.28 & 21.48\\
					\multicolumn{1}{|c|}{}&\text{E2E}	& 2.30 & 16.80 & 18.80\\
					\hline
				\end{tabular}
			\caption{\label{tab:asr_swbd} \ac{asr} performance in \ac{wer}(\%) on train, validation and test splits from \ac{swbd} and \ac{icsi}.}
			\vspace{-1em}
		\end{table}


\section{Experimental results}
\subsection{Experiments on \acl{mt}}
Table \ref{tab:nospkr} shows the results from a baseline model and our proposed model trained on \ac{mt} with context length varying from 1 to 5. The baseline model is a \ac{cnn} that receives as input an utterance at a time followed by a max pooling operation and a softmax layer. 

\begin{table}[h!]
	\centering
	\footnotesize
	\begin{tabular}{|c|r|r|}
		\hline
		\textbf{Context} & \multicolumn{1}{c|}{\textbf{\ac{icsi}}} & \multicolumn{1}{c|}{\textbf{\ac{swbd}}} \\ \hline
	0 (baseline) & 80.2  (80.4, 80.0)  &  72.0 (72.2, 71.6) \\\hline
		1 & 84.6 (84.6, 84.7) & 74.1 (73.2,	74.9) \\ \hline
		2 & {\bf84.7} (84.6, 84.7) & {\bf74.6} (74.5, 74.8) \\ \hline
		3 & 84.6 (84.5, 84.6) & 74.5 (74.2, 74.8) \\ \hline
		4 & {\bf84.7} (84.4, 84.8) & 74.1 (73.6, 74.6) \\ \hline
		5 & 84.6 (84.4, 84.8) & 74.2 (73.8, 74.5) \\ \hline
	\end{tabular}
	\caption{Baseline model and proposed model's accuracy (\%). For the latter we report for contexts from 1 to 5. Results appear like \textit{average (minimum, maximum)} calculated on 5 runs.}
	\label{tab:nospkr}
\end{table}

On average, for \ac{icsi} the best results were obtained with context 2 and 4 achieving 84.7\%, whereas for \ac{swbd} the model with context 2 achieves the highest performance, i.e. 74.6\%. 
To the best of our knowledge and under the setup in \cite{Lee2016}, these are state-of-the-art results on \ac{icsi} and \ac{swbd} outperforming \cite{Ortega2017}. For further experimentation in this paper, the context is fixed to 2.

\subsection{Experiments on \acl{at}}
We tested the pretrained models on \ac{at} from both \ac{asr} systems in order to see the impact on the accuracy (see Table \ref{tab:train_MT}). As expected, the performance dropped down dramatically due to the \ac{wer} and the lack of punctuation. On both datasets, the negative impact was higher when the model was tested on \ac{tdnn}/\ac{hmm} transcriptions. 

\begin{table}[h!]
	\centering
	\footnotesize
	\begin{tabular}{|c|r|r|}
		\hline
		\textbf{Transcriptions} & \multicolumn{1}{c|}{\textbf{\ac{icsi}}} & \multicolumn{1}{c|}{\textbf{\ac{swbd}}} \\ \hline
		\ac{mt} & {\bf84.7} (84.6, 84.7) & {\bf74.6} (74.5, 74.8) \\ \hline
		\ac{tdnn}/\ac{hmm} & 59.2 (58.9, 59.7) & 65.7	(65.4, 66.0) \\ \hline
		\ac{e2e} & 66.1 (65.7, 66.3) & 67.4 (66.6, 67.9) \\ \hline
	\end{tabular}
	\caption{Accuracy (\%) of the model trained on \ac{mt} with context 2  and tested on \ac{mt} and \ac{at}.}
	\label{tab:train_MT}
\end{table}

Afterwards, we retrained the DA model with \ac{at}. Tables \ref{tab:train_TDNN/HMM} and \ref{tab:train_ET} show the accuracy of training with \ac{tdnn}/\ac{hmm} and \ac{e2e} transcriptions, respectively.
Training on \ac{at} increases the accuracy when testing on \ac{at} and decreases it when testing \ac{mt} as expected.
In case of \ac{icsi}, the accuracy is slightly worse when training on \ac{at} from one system and testing on the other.
However in case of \ac{swbd}, the accuracy is always better when testing on the \ac{at} generated from the \ac{e2e} system.
Overall, we observed the best performance when training and testing on \ac{at} generated from the \ac{e2e} system on both datasets (76.6\% on \ac{swbd} and 68.7\% on \ac{icsi}. See Table \ref{tab:train_ET}).

\begin{table}[h!]
	\centering
	\footnotesize
	\begin{tabular}{|c|r|r|}
		\hline
		\textbf{Transcriptions} & \multicolumn{1}{c|}{\textbf{\ac{icsi}}} & \multicolumn{1}{c|}{\textbf{\ac{swbd}}} \\ \hline
		\ac{mt} & 64.2 (62.8, 65.7) & 66.9 (64.4, 69.5) \\ \hline
		\ac{tdnn}/\ac{hmm} & {\bf74.0} (73.9, 74.1) & 67.9	(67.5, 68.2) \\ \hline
		\ac{e2e} & 71.1 (70.8, 71.7) & {\bf68.6} (68.1, 68.8) \\ \hline
	\end{tabular}
	\caption{Accuracy (\%) of the model trained on \ac{tdnn}/\ac{hmm} transcriptions with context 2 and tested on \ac{mt} and \ac{at}.}
	\label{tab:train_TDNN/HMM}
\end{table}

\vspace{-1.1em}

\begin{table}[h!]
	\centering
	\footnotesize
	\begin{tabular}{|c|r|r|}
		\hline
		\textbf{Transcriptions} & \multicolumn{1}{c|}{\textbf{\ac{icsi}}} & \multicolumn{1}{c|}{\textbf{\ac{swbd}}} \\ \hline
		\ac{mt} & 70.9	(68.3,	72.7) & 66.6 (65.3, 70.0) \\ \hline
		\ac{tdnn}/\ac{hmm} & 73.2	(73.1,	73.3) & 67.1	(66.2, 67.6) \\ \hline
		\ac{e2e} & \textbf{76.6}	(76.5,	76.7) & \textbf{68.7} (68.4, 69.0) \\ \hline
	\end{tabular}
	\caption{Accuracy (\%) of the model trained on \ac{e2e} transcriptions with context 2  and tested on \ac{mt} and \ac{at}.}
	\label{tab:train_ET}
\end{table}

One of the main differences between \ac{mt} and \ac{at} is that the latter has no punctuation. In \cite{Ortega2018}, it was shown that punctuation provides strong lexical cues. Therefore, we retrained the model on \ac{icsi}'s \ac{mt} without punctuation. \ac{swbd} was not considered because the NXT-\ac{swbd} has no punctuation.


\begin{table}[h!]
	\centering
	\footnotesize
	\begin{tabular}{|c|c|c|}
		\hline
		\textbf{\ac{icsi}} & \bf With & \bf Without \\
		\textbf{transcriptions} & \bf punctuation & \bf punctuation \\ \hline
		\ac{mt} & {\bf84.7} (84.6, 84.7) & {\bf81.3} (81.1, 81.5) \\ \hline
		\ac{tdnn}/\ac{hmm} & 59.2 (58.9, 59.7) &  69.3 (69.3, 69.4) \\ \hline
		\ac{e2e} & 66.1 (65.7, 66.3) &  76.2 (76.0, 76.4) \\ \hline
	\end{tabular}
	\caption{Accuracy (\%) of the model with context $2$ trained on \ac{icsi}'s \ac{mt} without punctuation and tested on \ac{mt} and \ac{at}.}
	\label{tab:train_MT_nopunct}
\end{table}	

It can be seen from Table \ref{tab:train_MT_nopunct} that punctuation is a strong cue for \ac{da} classification. Nonetheless, it leads to a high negative impact while testing on AT without punctuation.
If \ac{mt} are used to train a model, it is advisable to remove punctuation. According to our results, by doing this a 10\% improvement in accuracy terms is achieved on both \ac{asr} transcriptions of \ac{icsi}.

\vspace{-1.1em}

\section{Conclusion}
\label{sec:conclusion}
We explored dialog act classification on \ac{mt} with a novel approach for context modeling that combines \acp{cnn} and \acp{crf}, reaching state-of-the-art results on two benchmark datasets (\ac{icsi} and \ac{swbd}).
We also investigated the impact of \ac{at} from two different \acl{asr} systems (hybrid \ac{tdnn}/\ac{hmm}  and \acl{e2e}) on the final performance. 
Experimental results showed that although the \acp{wer} are comparable, the \acl{e2e} \ac{asr} system might be more suitable for \acl{da} classification. Moreover, results confirm that punctuation yields central cues for the task suggesting that punctuation should be integrated into the \ac{asr} output in future works.


\clearpage
\bibliographystyle{IEEEbib}
\bibliography{ASR,DA_tagger}
\begin{acronym}[Bash]
    \acro{adagrad}[AdaGrad]{adaptive gradient algorithm}
    \acro{am}[AM]{attention mechanism}
    \acro{crf}[CRF]{conditional random field}
    \acro{cnn}[CNN]{convolutional neural network}
    \acro{da}[DA]{dialog act}
    \acro{dl}[DL]{deep learning}
    \acro{e2e}[E2E]{End-to-End}
    \acro{gd}[SGD]{stochastic gradient descent}
    \acro{icsi}[MRDA]{ICSI Meeting Recorder Dialog Act Corpus}
    \acro{hmm}[HMM]{hidden Markov model}
    \acro{nlp}[NLP]{natural language processing}
    \acro{rnn}[RNN]{recurrent neural network}
    \acro{svm}[SVM]{support vector machine}
    \acro{swbd}[SwDA]{Switchboard Dialog Act Corpus}
    \acro{asr}[ASR]{automatic speech recognition}
    \acro{nn}[NN]{neural network}
    \acro{mt}[MTs]{manual transcriptions}
    \acro{at}[ATs]{automatic transcriptions}
    \acro{e2e}[E2E]{End-to-End}
    \acro{cd}[CD]{context-dependent}
    \acro{tdnn}[TDNN]{time-delay neural network} 
    \acro{ctc}[CTC]{connectionist temporal classification}
    \acro{wer}[WER]{word error rate}
    \acro{mfcc}[MFCC]{Mel-frequency cepstral coefficient}
    \acro{gmm}[GMM]{Gaussian Mixture Model}
    \acro{espnet}[ESPnet]{End-to-End Speech Processing Toolkit}
\end{acronym}

\end{document}